\documentclass[a4paper,fleqn]{cas-dc}

\usepackage{caption}
\usepackage{subcaption}
\usepackage[numbers]{natbib}
\usepackage{hhline}
\usepackage{colortbl}
\usepackage{microtype}

\newcommand{\bigCI}{\mathrel{\text{\scalebox{1.07}{$\perp\mkern-10mu\perp$}}}}

\def\tsc#1{\csdef{#1}{\textsc{\lowercase{#1}}\xspace}}
\tsc{WGM}
\tsc{QE}
\tsc{EP}
\tsc{PMS}
\tsc{BEC}
\tsc{DE}

\begin{document}

\let\WriteBookmarks\relax
\def\floatpagepagefraction{1}
\def\textpagefraction{.001}
\shorttitle{Real-time EEG-ER using DWT on Full and Reduced Channel Signals}
\shortauthors{Bajada and Borg Bonello.}

\title [mode = title]{Real-time EEG-based Emotion Recognition using Discrete Wavelet Transforms on Full and Reduced Channel Signals}                     



\author[um]{Josef Bajada}[orcid=0000-0002-8274-6177]
\cormark[1]
\cortext[cor1]{Principal corresponding author}
\ead{josef.bajada@um.edu.mt}
\credit{Supervision, Conceptualization, Methodology, Software, Writing - Review \& Editing}

\author[um]{Francesco {Borg Bonello}}
\ead{francesco.borg.18@um.edu.mt}
\credit{Methodology, Investigation, Software, Writing - Original draft preparation}

\address[um]{Department of Artificial Intelligence, Faculty of ICT, University of Malta, Malta}

\begin{abstract}
Real-time EEG-based Emotion Recognition (EEG-ER) with consumer-grade EEG devices involves classification of emotions using a reduced number of channels. These devices typically provide only four or five channels, unlike the high number of channels (32 or more) typically used in most current state-of-the-art research. In this work we propose to use Discrete Wavelet Transforms (DWT) to extract time-frequency domain features, and we use time-windows of a few seconds to perform EEG-ER classification. This technique can be used in real-time, as opposed to post-hoc on the full session data. We also apply baseline removal preprocessing, developed in prior research, to our proposed DWT Entropy and Energy features, which improves classification accuracy significantly. We consider two different classifier architectures, a 3D Convolutional Neural Network (3D CNN) and a Support Vector Machine (SVM). We evaluate both models on subject-independent and subject dependent setups to classify the Valence and Arousal dimensions of an individual's emotional state. We test them on both the full 32-channel data provided by the DEAP dataset, and also a reduced 5-channel extract of the same dataset. The SVM model performs best on all the presented scenarios, achieving an accuracy of 95.32\% on Valence and 95.68\% on Arousal for the full 32-channel subject-dependent case, beating prior real-time EEG-ER subject-dependent benchmarks. On the subject-independent case an accuracy of 80.70\% on Valence and 81.41\% on Arousal was also obtained. Reducing the input data to 5 channels only degrades the accuracy by an average of 3.54\% across all scenarios, making this model appropriate for use with more accessible low-end EEG devices.
\end{abstract}


\maketitle

\section{Introduction}

EEG-based Emotion Recognition (EEG-ER) involves the inference of an individual's emotional state from Electroencephalogram (EEG) signals. There are numerous applications for EEG-ER, including emotion-adaptive Human Computer Interaction (HCI) \cite{brave2007emotion} and entertainment systems, that typically fall under the category of Affective Computing \cite{picard1999affective}. EEG-ER also has useful medical applications, such as diagnosing mood disorders \cite{keshky2018emotion} and tools that could assist in psychiatric therapy. 

\begin{figure}
    \centering
    \includegraphics[width=0.45\textwidth]{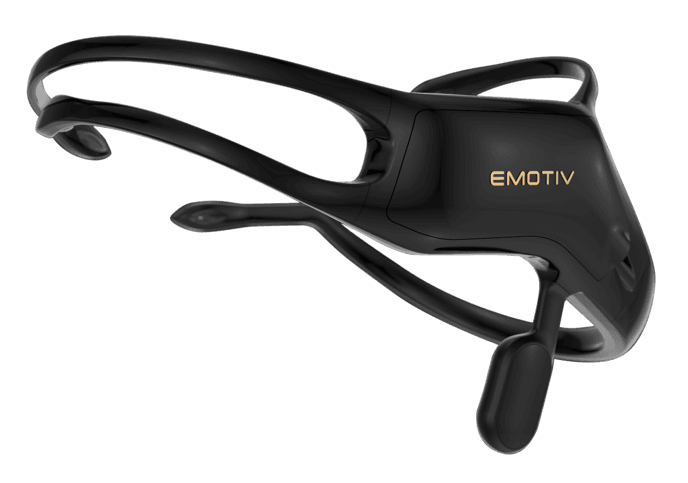}
    \caption{The 5-channel Emotiv Insight EEG Device}
    \label{fig:emotivinsight}
\end{figure}

EEG devices are non-invasive and read brain activity in real-time
\cite{Niemic2004StudiesEmotion}. They come in various shapes and sizes, ranging from expensive medical-grade 32 or 62 channel caps, to headbands with 5 channels or less targeting the consumer market. One of these more affordable devices is the Emotiv Insight, shown in Figure \ref{fig:emotivinsight}, which has 5 channels together with 2 reference channels. This brand is quite popular with researchers looking into Brain Computer Interface (BCI) technology \cite{MartinStrmiska2018MeasuringDevices,duvinage2013performance,stytsenko2011evaluation}, and the cheaper models with limited channels are easier to wear and more practical for day-to-day use in BCI applications.

Most of the current research in EEG-ER makes use of the full 32 or 62 channels \cite{TorresEEG-BasedSurvey}. These are often based on either the Database for Emotion Analysis using Physiological signals (DEAP) \cite{KoelstraDEAP:Signals} dataset (32 channels) or the SJTU Emotion EEG Dataset (SEED) \cite{duan2013differential,Zheng2015InvestigatingNetworks} (62 channels) \cite{Alarcao2019EmotionsSurvey,cimtay2020investigating,fernandez2021cross,jirayucharoensak2014eeg,li2020latent,ShiblyMokatren2020EEGConfiguration,Tao2020EEG-basedAttention,TorresEEG-BasedSurvey,Yang2018ContinuousRecognition}. Other less popularly used datasets are also available, such as the DREAMER \cite{katsigiannis2017dreamer} dataset (14 channels) and the MAHNOB-HCI \cite{soleymani2011multimodal} (32 channels).

In this research, we propose techniques that improve on the current approaches for real-time EEG-ER found in literature. The proposed models are based on preprocessing the EEG signals using Discrete Wavelet Transforms, which are often used to perform time-frequency analysis of nonstationary signals, such as EEG. As part of our preprocessing, we also remove the baseline reference signal (provided as part of the DEAP \cite{KoelstraDEAP:Signals} dataset), to normalise the signal, yielding a higher classification accuracy. We consider both the subject independent case, where a single model is used across all individuals, and also the subject dependent case, where a separate model is trained for each individual. Our best 32-channel model achieves a high accuracy that exceeds the current state-of-the-art subject-dependent models that also use the DEAP \cite{KoelstraDEAP:Signals} dataset. We also propose an adaptation of our model to make it work with 5 channels, corresponding to those used by low-end devices such as the Emotiv Insight, while still retaining a high accuracy.

\subsection{Contributions}
The contributions of this work are as follows:
\begin{enumerate}
    \item A novel approach to preprocess 32-channel EEG signals for emotion recognition using Discrete Wavelet Transforms combined with baseline removal.
    \item A real-time emotion classification model that yields an accuracy that exceeds 95\% for subject-dependent and 83\% for subject-independent models.
    \item An adaptation of the proposed model from 32 channels to 5 channels that retains a high subject-dependent accuracy of 92\%, together with 81\% for the subject-independent case.
\end{enumerate}

This study proposes a set of generic high accuracy classification models for emotion recognition using both 32-channel and 5-channel setups, for both subject-independent and subject dependent scenarios, that can be used in real-time. 

\section{Background}
\label{eeg-er_background}

Emotion recognition has been attempted using a variety of different inputs such as speech, facial gestures and other physiological measures \cite{ackermann2016eeg}. Using EEG signals for emotion recognition has been seeing an increase in popularity and research interest due to it being less susceptible to manipulation by the individual \cite{TorresEEG-BasedSurvey}. In this section we define how we are going to approach real-time emotion classification, together with a review of the state-of-the-art EEG-ER models found in existent literature to date. 

\subsection{Emotion Classification}

\begin{figure}[h!]
    \centering
    \includegraphics[width=0.45\textwidth]{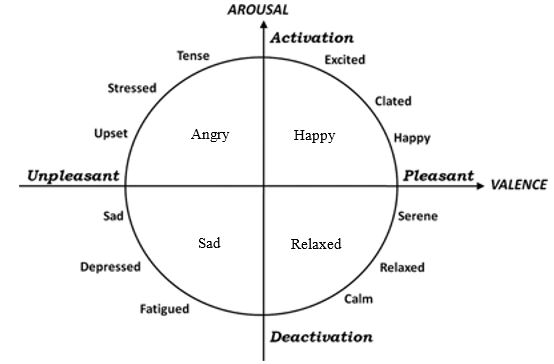}
    \caption{Russell's Circumplex Model of Affect. \cite{Russell1980AAffect}}
    \label{fig:russelmodel}
\end{figure}

One of the most popular models for emotion classification is the Russell's Circumplex Model of Affect \cite{Russell1980AAffect}. It maps different emotions on a 2-dimensional space, with Valence on the x-axis, and Arousal on the y-axis. A High Valence (HV) corresponds to a pleasant emotion, while a Low Valence (LV) corresponds to an unpleasant one. Similarly, a High Arousal (HA) corresponds to an active emotion (a high level of excitement) while a Low Arousal (LA) corresponds to an inactive one. Each quadrant is associated with four categories of emotions being Angry (HALV), Happy (HAHV), Sad (LALV), and Relaxed (LAHV). Twelve individual emotions, such as Tense, Calm and Fatigued, are then mapped across these four quadrants, as shown in Figure \ref{fig:russelmodel}.

The DEAP dataset \cite{KoelstraDEAP:Signals} is often used in EEG-ER benchmarks because it provides ratings for both Valence and Arousal, which correspond to the Russell Model \cite{Russell1980AAffect}. It consists of EEG signals recorded from thirty-two participants aged between 19 and 37. Each participant was asked to watch forty 60-second long music videos during which the EEG signals were recorded. Prior to each of the 60-second trials, a 3-second pre-trial baseline was also recorded to measure the \textit{rest state} of the subject prior to evoking an emotion through the music videos. After watching each video the participant filled out a Self-Assessment Manikin (SAM) \cite{bradley1994measuring} where they rated Valence and Arousal on a continuous scale from 1-9. 

\subsection{Real-Time EEG-ER}

To perform EEG-ER in real-time, EEG signals are typically segmented into samples corresponding to time windows of a few seconds duration \cite{Rozgic2013RobustFusion}. Various characteristics of the signals can then extracted from these samples, such as Spectral Power, Kurtosis coefficients, Band Power, and Hjorth parameters \cite{Atkinson2016ImprovingClassifiers}. When these features were used with Support Vector Machines (SVM) using a Radial Basis Function (RBF) kernel a mean accuracy of 73.1\% on both the Valence and Arousal dimensions was achieved \cite{Atkinson2016ImprovingClassifiers}, across all participants from the DEAP dataset. However, these features only carry information from the time-domain. Since EEG signals are non-stationary, techniques that extract time-frequency domain features are more adequate \cite{Rosso2001WaveletSignals}.

A Discrete Wavelet Transform (DWT) \cite{daubechies1992ten} is a transform that processes a signal in the time-frequency domain. It breaks down the signal into a number of components, each  corresponding to a time series of coefficients that describe how the signal changes over time within a frequency band. When different wavelets were evaluated, the features extracted using \textit{db4} wavelet achieved the best results compared to other feature sets computed using other wavelets \cite{Feradov2020EvaluationSignals}. When the \textit{db4} wavelet was used to extract features across 4-second time windows, an accuracy of 86.8\% on Valence and 84.1\% on Arousal was achieved \cite{Mohammadi2017Wavelet-basedSignal}. 

A 5-level DWT-\textit{db4} was also used to compute the wavelet entropy and energy features across 2-second time frames \cite{Ali2016EEG-basedApplications}. These features were used with SVM-RBF, outperforming Quadratic Discriminant Analysis (QDA) and k-Nearest Neighbours (kNN) for subject-independent EEG-ER. The performance of SVM-RBF was also compared to that of Artificial Neural Networks (ANNs), where SVM-RBF also had a superior accuracy on the DEAP dataset \cite{BazgirEmotionSignals}.

A 4-level DWT-\textit{db4}, computing wavelet entropy and energy across theta, alpha, beta and gamma sub bands, was also used on an SVM-RBF \cite{ShiblyMokatren2020EEGConfiguration}, achieving a cross-validated accuracy of 83.4\% on Valence and 81.4\% on Accuracy for the subject-independent case, rising to 85.3\% and 83.0\% respectively for the subject-dependent case. 

Although traditional machine learning models such as SVMs are able to classify Valence and Arousal with high accuracy, such models do not take into account spatial information \cite{Li2017EmotionNetwork} such as the topological relationship between the different channels and their position on the skull. An approach inspired from computer vision transforms EEG samples into grid-like frames, that include channel and scale information on each axis \cite{Li2017EmotionNetwork}. A Convolution Neural Network (CNN) is combined with a Recurrent Neural Network (RNN) forming a Convolutional Recurrent Neural Network (CRNN). This achieved a mean subject-dependent accuracy of 72.1\% on Valence and 74.1\% on Arousal. 

A 3D CNN model was also proposed, which makes use of a 3-dimensional frame consisting of a stack of grids \cite{Salama2018EEG-basedNetworks,Yang2018ContinuousRecognition}. In the case of \citet{Salama2018EEG-basedNetworks}, with each frame consists of the readings from each channel for the duration of the set time-window. This is preprocessed using a high pass filter and  band stop filter to reduce noise, which are then fed into a CNN, achieving an accuracy of 87.44\% for Valence and 88.49\% for Arousal. In the case of \citet{Yang2018ContinuousRecognition}, each value in the 3D frame corresponds to the Differential Entropy (DE) for each channel, positioned according to the channel's sensor location on the scalp. Furthermore, the initial 3-second pre-trial baseline included in the DEAP dataset, which corresponds to the subject's \textit{rest} state, was used to normalise the readings. A significant increase in subject-dependent accuracy on both Valence and Arousal was observed when this baseline removal preprocessing was applied, with Valence accuracy increasing from 68.6\% to 89.5\% and Arousal increasing from 69.6\% to 90.2\%. This baseline preprocessing technique was also used with an Attention-based CRNN (ACRNN) model \cite{Tao2020EEG-basedAttention}, achieving a mean subject-dependent accuracy of 92.7\% for Valence and 93.1\%. 

All of the aforementioned models make use of the full 32 channel data provided by the DEAP dataset \cite{KoelstraDEAP:Signals}. For these models to be used in real life, high-end devices are needed, which are not practical for consumer applications. A 5-channel approach was used with channels, $F3$, $CP5$, $FP2$, $FZ$, and $FC2$, chosen for their position on the prefrontal region \cite{jie2014emotion}. A sample entropy technique was used to extract features, which were then used to train an SVM. This model achieved an accuracy of 80.43\% on Valence and 71.16\% on Arousal. 

\begin{figure}[h!]
    \centering
    \includegraphics[width=0.45\textwidth]{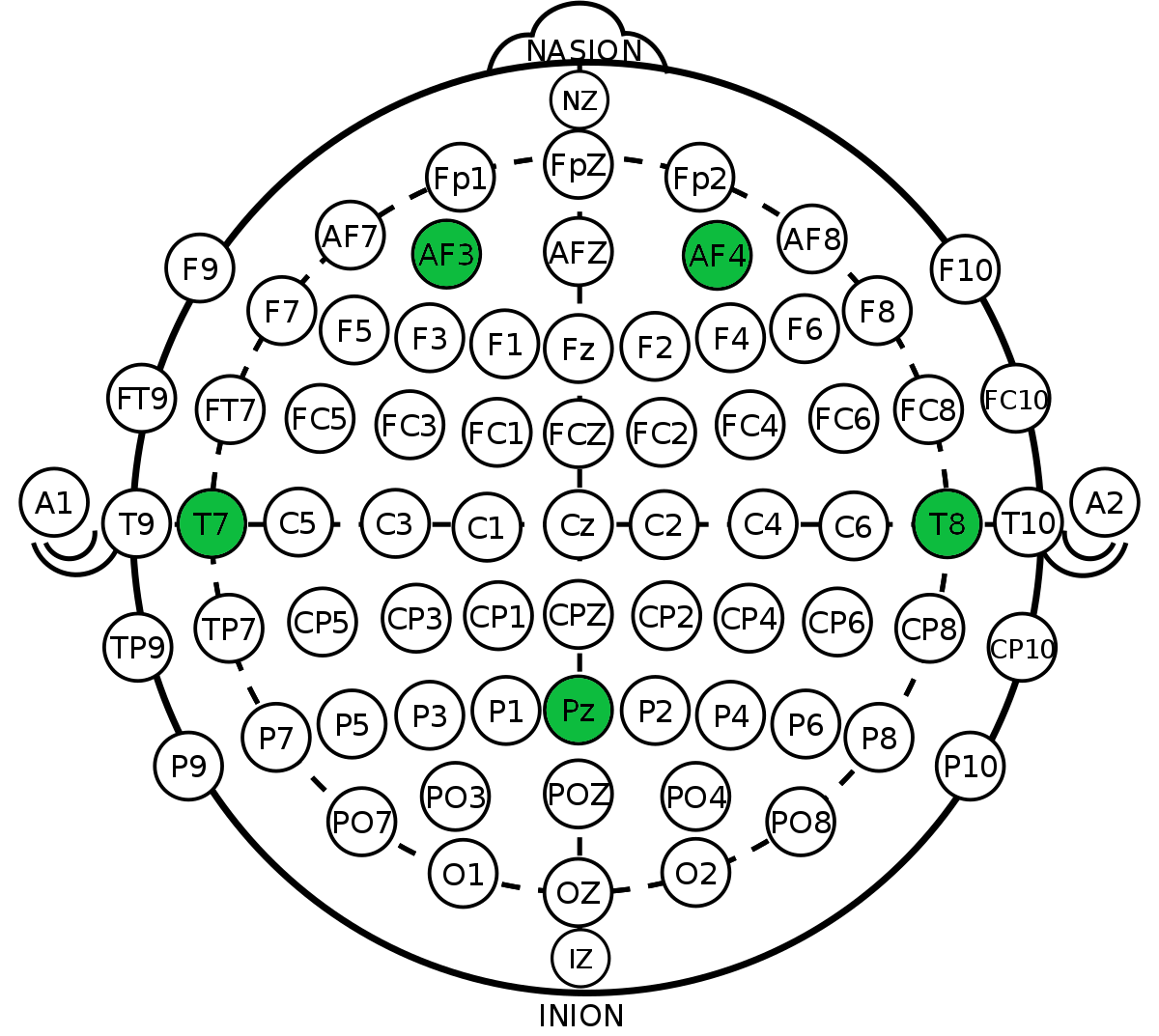}
    \caption{The selected sensors for the reduced 5-channel scenario mapped onto the International 10-20 system.}
    \label{fig:1020sensors}
\end{figure}

In this research we will make use of DWT to extract time-frequency domain features, and use them with the state-of-the-art Machine Learning models, SVM and 3DCNN, to determine which perform best in both the full 32-channel and also in the reduced 5-channel scenarios. The chosen for the latter are $AF3$, $T7$, $Pz$, $AF4$ and $T8$, highlighted in Figure \ref{fig:1020sensors}, corresponding to those used by the Emotiv Insight EEG device. Both subject-independent and subject-dependent flavours will be compared. We will also make use of the DEAP dataset \cite{KoelstraDEAP:Signals}, and analyse the difference in accuracy when using baseline removal preprocessing \cite{Yang2018ContinuousRecognition,Tao2020EEG-basedAttention}.

\section{Methodology}

We model the Emotion Classification problem as two separate binary classifiers, one for Valence (Low vs High) and one for Arousal (Low vs High). The output of both classifiers can then combined to obtain the emotion category of one of the four quadrants of Russell's Circumplex Model of Affect \cite{Russell1980AAffect} shown in Figure \ref{fig:russelmodel}. 

\subsection{Data Preprocessing}

Since the DEAP \cite{KoelstraDEAP:Signals} dataset provides ratings between 1 and 9 for both Valence and Arousal, ratings were divided into the two classes, High and Low, with a threshold of 5, consistent with previous work \cite{Yang2018ContinuousRecognition,Tao2020EEG-basedAttention}. 

The DEAP \cite{KoelstraDEAP:Signals} dataset is available in both raw and preprocessed forms. The raw signals are at a frequency of 512Hz. The preprocessed signals were downsampled to 128Hz as well as applying a 4.0-45.0Hz bandwidth filter. The EOG components, representing the electrical activity of the eyeball and eyelid motions, were also removed. The data from each channel was then processed using  \textit{common average referencing}, where the average of all electrical activity is subtracted from each signal in the recording. For this reason, the preprocessed version could only be used for the 32-channel study.  For the 5-channel case, the raw data was imported into MNE\footnote{ \url{https://mne.tools/stable/index.html.}}, retaining the data from $AF3$, $T7$, $Pz$, $AF4$ and $T8$, and preprocessed in the same way. 

\begin{figure}
    \centering
    \includegraphics[width=0.48\textwidth]{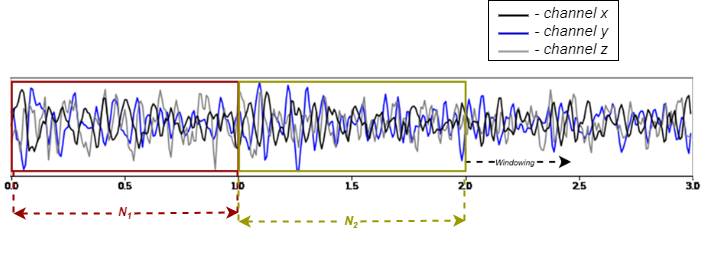}
    \caption{Time-windows of length 1 second.}
    \label{fig:timewindow}
\end{figure}

Each EEG recording is 63 seconds in length. We therefore need to segment the data into shorter time frames in order to simulate real-time classification. In our study we consider time-windows of length $\tau \in \{1,3\}$, which allow us to also make use of baseline removal preprocessing on each time window. We use a tumbling window approach, where each frame is non-overlapping, as shown in Figure \ref{fig:timewindow}.  

\subsection{Feature Extraction}
\label{sec:features}
We extract features from the time-frequency domain using DWT, which is often used to extract features from nonstationary signals \cite{wang2011discrete} such as EEG \cite{Feradov2020EvaluationSignals,Kumar2017WaveletANN,thammasan2017familiarity,ozerdem2017emotion}. For each channel within each EEG sample we apply 4-level DWT using the \textit{db4} wavelet, deriving the coefficients for theta; alpha; beta; and gamma frequency bands. These coefficients then allow the calculation of entropy and energy as shown in Equations \eqref{waventropy} and \eqref{wavenergy}:

\begin{equation}
\label{waventropy}
\textup{ent}(x) = -\sum_{t=1}^{n} x(t)^{2}\log(x(t)^{2})
\end{equation}

\begin{equation}
\label{wavenergy}
\textup{eng}(x) = \sum_{t=1}^{n} x(t)^{2}
\end{equation}

where $x$ is the signal in question and $n$ is the coefficient value at timestep $t$ \cite{Mohammadi2017Wavelet-basedSignal,ShiblyMokatren2020EEGConfiguration, Yu2021EmotionAdaBoost.}. These calculations are carried out using  PyWavelets\footnote{\url{https://github.com/PyWavelets/pywt}}  \cite{LeePyWavelets:Analysis} and EBAPy\footnote{ \url{https://github.com/DustinCarrion/EBAPy}} \cite{Carrion-Ojeda2021EBAPy:Applications}. 

\subsubsection{Baseline Removal}
Each EEG recording in the DEAP \cite{KoelstraDEAP:Signals} dataset begins with a 3-second pre-trial baseline period, during which the individual was not shown any stimulus, thus measuring the emotional \textit{rest state}. While this was previously done for DE \cite{Yang2018ContinuousRecognition,Tao2020EEG-basedAttention}, we now adapt the same technique for DWT-based wavelet entropy. 

When using baseline removal preprocessing, the features are modified to reflect the difference between the rest state and active state. After the features are extracted for both baseline and active period, the difference between both is calculated across the full 60-second sample. We consider two time-window sizes, $\tau = 1$ second \cite{Yang2018ContinuousRecognition} and $\tau = 3$ seconds. The baseline feature is calculated from the mean across all baseline segments, $K$, as shown in Equation \eqref{baseline_calc1}:

\begin{equation}
\label{baseline_calc1}
f_{c,s}^{b} = e_{c,s}^{b} - \frac{1}{K} \sum_{k=1}^{K} r_{c,k}^{b}
\end{equation}

where $f_{c,s}^{b}$ is the computed feature for frequency band $b$, channel $c$ and segment $s$. $e_{c,s}^{b}$ is the evoked feature for the same band, channel and segment, and $r_{c,k}^{b}$ is the baseline feature at rest for the same band and channel, and baseline segment $k$. $K$ corresponds to the total baseline segments. Since the DEAP \cite{KoelstraDEAP:Signals} dataset has a 3-second baseline, when the time-window is of size $\tau = 1$ second, the number of baseline segments, $K$, is $3$. On the other hand, when the time-window is of size $\tau = 3$ seconds, the base line segment is the whole reference baseline included in the dataset, and thus $K = 1$. 

\subsection{Classification}

We consider two types of classifiers, a 3D CNN, similar to the one described by \cite{Yang2018ContinuousRecognition}, and an SVM with RBF kernel similar to the one used in \cite{Mohammadi2017Wavelet-basedSignal,ShiblyMokatren2020EEGConfiguration}. 

The 3D CNN approach was chosen since it preserves the topological relationship of the readings as positioned on the subject's scalp. We use the DWT Entropy, as described in Equation \ref{wav_entropy}, and also evaluate a stacked model where both DWT Entropy and Differential Entropy \cite{Yang2018ContinuousRecognition} are combined together. 

The SVM was chosen due to its simple architecture, making it a very good candidate for real-time use, together with its high classification accuracy as reported in recent work \cite{Mohammadi2017Wavelet-basedSignal,ShiblyMokatren2020EEGConfiguration}. In our case, the input features for the SVM are the DWT Entropy, as described in Equation \ref{wav_entropy} and DWT Energy, as described in Equation \ref{wav_energy}.

For both types of classifiers, we measure the difference in accuracy when applying the baseline removal technique on all models, for both subject independent and subject dependent scenarios. 

The models were trained using a CUDA-compatible GPU, NVIDIA RTX 2060 SUPER 8GB. All samples were shuffled and a stratified n-fold cross validation using the Scikit-learn\footnote{ \url{http://scikit-learn.org/}} library to evaluate the performance of the proposed model in both subject-dependent and subject-independent cases. 

\subsubsection{3DCNN with DWT Entropy}
\label{sec:3dcnn}

We adapt the 3DCNN model \cite{Yang2018ContinuousRecognition} to use the DWT entropy instead of DE. It considers a stack of $9\times9$ grids containing channel feature values located according to their corresponding topological position on the scalp. 

\begin{figure*}
    \centering
    \includegraphics[width=0.85\textwidth]{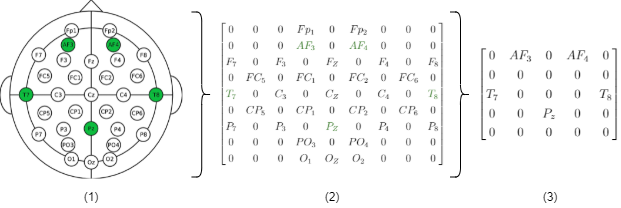}
    \caption{(1) - the positions of 32 electrodes, with the selected 5 channels highlighted in green; (2) - the matrix representation of all 32 electrodes; (3) - the compact matrix representation of the selected 5 channels. (Image adapted from \cite{Yang2018ContinuousRecognition}.)}
    \label{5x5grid}
\end{figure*}

For the reduced channel version, we reduce the size of the grid to a $5\times5$ matrix, in order to eliminate the unwanted channels while retaining the same topology, as shown in Figure \ref{5x5grid}. This process reduces the sparsity of the matrix, while retaining the same inter-channel topological relationship. 

\begin{figure}
    \centering
    \includegraphics[width=0.48\textwidth]{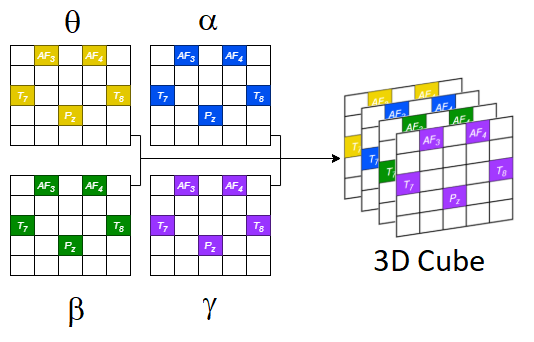}
    \caption{Four 2D grids, one for each sub-band, are combined together into a single 3D cube. The feature values are placed in their corresponding channel location. for the 5-channel case each grid will be a $5\times5$ matrix (as shown), while for the 32-channel case each grid will be a $9\times9$ matrix.}
    \label{3dcnn_cube}
\end{figure}

Four of these grids are stacked on top of each other forming a 3D EEG cube, with each grid representing a specific frequency band: theta ($\theta$); alpha ($\alpha$); beta ($\beta$); and gamma ($\gamma$), as shown in Figure \ref{3dcnn_cube}. The formation of the 3D cube is similar to a colour image in a computer vision context; where the DWT entropy value for each of the frequency bands ($\theta$, $\alpha$, $\beta$, $\gamma$) is used instead of the colour intensity of the 3 colour channels (Red, Green, Blue). 

\begin{figure*}
    \centering
    \includegraphics[width=0.8\textwidth]{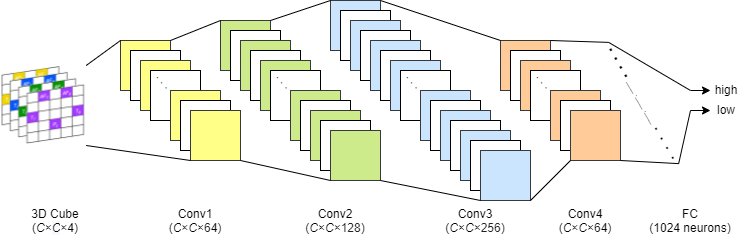}
    \caption{Continuous Convolutional Neural Network with 3D DWT entropy input. $C \in \{5,9\}$ represents the number of rows and columns of the input grids. Conv$n$ layers are 3D Convolutional Layers, while FC is a Fully Connected dense layer of 1024 neurons. (Image adapted from \cite{Yang2018ContinuousRecognition}.)}
    \label{3dcnn}
\end{figure*}

The 3D CNN architecture is identical to the original model \cite{Yang2018ContinuousRecognition}, shown in Figure \ref{3dcnn}. This CNN structure is continuous since it does not use pooling layers between the convolutional layers. This is because the input sample is far smaller than the typical input used in computer vision, and therefore a pooling operation is not necessary.  The kernel size is set to 4 with a stride of 1. The first convolutional layer consists of 64 feature maps; doubling the amount in the next two layers with 128 in the second and 256 in the third. The fourth layer fuses the feature maps back into 64 $9\times9$ or $5\times5$ feature maps, which are then passed to a fully connected layer of $1024$ neurons that map them to a feature vector $f \in R^{1024}$. This is finally passed to a softmax layer to obtain the High or Low prediction \cite{Yang2018ContinuousRecognition}. 

The TensorFlow\footnote{\url{https://www.tensorflow.org/}} framework was used to implement and train the 3DCNN models. The Adam optimizer was used to minimize the cross-entropy loss, with the initial learning rate set to $10^{-4}$. In order to avoid overfitting and improve the overall generalization capability of the system, the dropout probability was set to 0.5, and L2 regularization was also added with a penalty strength of 0.5. We applied a stratified 10-fold cross-validation to both the subject-independent and subject-dependent cases, to minimise overfitting bias from the accuracy results. 

\subsubsection{3DCNN with Stacked DWT Entropy and DE}

In the previous section, we made use of only the DWT entropy for the input features that form the 3D EEG cubes. Since DE also performed considerably well \cite{Yang2018ContinuousRecognition}, we also experiment with combining both DWT entropy and DE into an 8-layered EEG cube, with the layers 1 to 4 consisting of the DWT entropy values and layers 5 to 8 corresponding to DE. The architecture of the model was kept identical, as described in Section \ref{sec:3dcnn}. The same Tensorflow configuration was used, together with a stratified 10-fold cross-validation for both subject independent and dependent cases.

\subsubsection{SVM with RBF kernel.}
\label{sec:svmrbf}

\begin{table*}
\centering
\resizebox{\textwidth}{!}{%
\begin{tabular}{cc}
5 channels & 32 channels \\
\begin{tabular}{|c|c|c|c|c|c|}
\hline
\multicolumn{2}{|c|}{\textbf{Model}} & \multicolumn{4}{c|}{\textbf{Hyperparameters}} \\ \hline
\textbf{Size} & \textbf{Baseline} & \textbf{VAL} & \textbf{ARO} & \textbf{VALARO} & \textbf{AROVAL} \\ \hline
1 & No & \textit{C}:1, \textit{g}:1 & \textit{C}:1, \textit{g}:1 & \textit{C}:1, \textit{g}:1 & \textit{C}:1, \textit{g}:1 \\ \hline
1 & Yes & \textit{C}:50, \textit{g}:1 & \textit{C}:50, \textit{g}:1 & \textit{C}:50, \textit{g}:1 & \textit{C}:50, \textit{g}:1 \\ \hline
3 & No & \textit{C}:1, \textit{g}:1 & \textit{C}:1, \textit{g}:1 & \textit{C}:1, \textit{g}:1 & \textit{C}:1, \textit{g}:1 \\ \hline
3 & Yes & \textit{C}:50, \textit{g}:1 & \textit{C}:50, \textit{g}:1 & \textit{C}:50, \textit{g}:1 & \textit{C}:50, \textit{g}:1 \\ \hline
\end{tabular} & %

\begin{tabular}{|c|c|c|c|c|c|}
\hline
\multicolumn{2}{|c|}{\textbf{Model}} & \multicolumn{4}{c|}{\textbf{Hyperparameters}} \\ \hline
\textbf{Size} & \textbf{Baseline} & \multicolumn{1}{c|}{\textbf{VAL}} & \multicolumn{1}{c|}{\textbf{ARO}} & \multicolumn{1}{c|}{\textbf{VALARO}} & \multicolumn{1}{c|}{\textbf{AROVAL}} \\ \hline
1 & No & \textit{C}:1, \textit{g}:1 & \textit{C}:1, \textit{g}:1 & \textit{C}:1, \textit{g}:1 & \textit{C}:1, \textit{g}:1 \\ \hline
1 & Yes & \textit{C}:50, \textit{g}:1 & \textit{C}:50, \textit{g}:1 & \textit{C}:50, \textit{g}:1 & \textit{C}:300, \textit{g}:1 \\ \hline
3 & No & \textit{C}:50, \textit{g}:1 & \textit{C}:300, \textit{g}:0.001 & \textit{C}:1, \textit{g}:1 & \textit{C}:1, \textit{g}:1 \\ \hline
3 & Yes & \textit{C}:50, \textit{g}:1 & \textit{C}:200, \textit{g}:1 & \textit{C}:50, \textit{g}:1 & \textit{C}:50, \textit{g}:1 \\ \hline
\end{tabular} \\ %
\end{tabular}}
\caption{Optimal SVM RBF hyperparameters for subject-independent models found via grid search.}
\label{tab:independentHyperparams}
\end{table*}

Two SVM classifiers with RBF kernel were implemented, one for Valence and one for Arousal. Each SVM took an input feature vector of size $\Omega \times B \times F$, where $B = 4$ corresponds to the number of frequency bands ($\theta$, $\alpha$, $\beta$, $\gamma$), $F = 2$ represents the number of extracted features per band (wavelet energy, $eng$, and entropy, $ent$), and $\Omega \in \{5,32\}$ corresponds to the amount of channels. The sizes of the feature vectors for the full 32-channel and reduced 5-channel cases are 256 and 40 respectively. 

 The models were implemented using ThunderSVM\footnote{\url{https://github.com/Xtra-Computing/thundersvm25}} to utilize the GPU. Grid Search with Cross Validation was used to find the best hyperparameter values for $\mathtt{C}$ and $g$ for both subject-independent and subject-dependent cases. Table \ref{tab:svmHyperparameters} shows the search space of possible parameters considered. 
 
  \begin{table}
     \centering
     \begin{tabular}{|c|c|}
     \hline
     Parameter & Values \\
     \hline
        $\mathtt{C}$  & 1, 50, 100, 200, 300 \\
        $g$  & 0.00001, 0.001, 1, 50, 100 \\ 
    \hline
     \end{tabular}
     \caption{Hyperparameter values considered for SVM with RBF Kernel.}
     \label{tab:svmHyperparameters}
 \end{table}
 
 For the subject-independent case, a sample of one third the size of the full dataset was taken, while retaining the same class distribution of the entire set. From this sample, we found the optimal hyperparameters based on a stratified 6-fold cross validation. These were computed for each model and are presented in Table \ref{tab:independentHyperparams}. 
 
 Since it is impractical to optimise the hyperparameters for every individual, we set out to find one generic hyperparameter value for $\mathtt{C}$ that performs well across all subjects. We first ran a Grid Search on each individual subject and then computed the mean across all individuals, rounding it to the nearest value in the search space of values in Table \ref{tab:svmHyperparameters}, resulting in $\mathtt{C} = 200$. The $g$ parameter was left to the default setting used by the ThunderSVM implementation, which scales it with respect to the number of features and variance of the input data, $g = 1  / (features \times variance)$, thus adapting to each subject's data automatically. This configuration was then used to train the subject-dependent models for each participant using a stratified 8-fold cross validation.

\subsubsection{Chained SVM Classifiers}

Since we are using two separate classifiers for the Valence and Arousal, the models we have discussed so far assume that both classifiers are independent of each other, even though they are using the same input data. However, since the data is the same, these two dimensions are only \textit{conditionally independent} on the same emotional state represented by the data $(Arousal \bigCI  Valence | Emotion)$. This aspect was also observed in previous work \cite{jie2014emotion}, where Arousal was kept constant when classifying Valence, and vice versa. Discovering either Valence or Arousal can thus be used as an extra input to feed to the other classifier, helping it to get a more accurate prediction of the emotion given the same input data. 

The first model, $\mathit{Valence} | \mathit{Arousal}$ (VALARO), predicts the Arousal using the SVM described in Section \ref{sec:svmrbf}, and then feeds its output to the second SVM to predict the Valence given the known Arousal. The input feature vector of the second SVM is of size $(\Omega \times B \times F) + 1$, where the last feature is a one-hot encoding of the Arousal (LA = 0, HA = 1). 

The second model, $\mathit{Arousal} | \mathit{Valence}$ (AROVAL), first predicts the Valence, and then feeds the prediction to the second SVM to predict the Arousal given the known Valence. The input feature vector of the second SVM is also of size $(\Omega \times B \times F) + 1$, where the last feature is a one-hot encoding of the Valence (LV = 0, HV = 1).

The two models cannot be used together, but only one of them can be used in conjunction with its SVM counterpart that predicts the first dimension, Arousal or Valence from just the input data. 

The hyperparameters for the subject independent models are listed in Table \ref{tab:independentHyperparams}, while the subject dependent models use the same $\mathtt{C} = 200$ and scaling $g$ configuration. 

\section{Results}
\label{results}

In this section we present results achieved for the proposed models using the DEAP \cite{KoelstraDEAP:Signals} dataset. We first analyse the data itself for correlation between each of the chosen 5 channels and the rest of the channels, to determine whether these channels are likely to carry enough information to perform accurate EEG-ER classification. 

The 5 proposed models, 3DCNN with DWT Entropy, 3DCNN with Stacked DWT Entropy and DE, SVM with RBF kernel using DWT Entropy and Energy, and the two Chained SVM models, VALARO and AROVAL are then evaluated. We test each model on two time-window sizes, 1 and 3 seconds, with and without baseline, for both subject-independent and subject-dependent cases on the full 32 channels and also the reduced 5-channel setup. 

\subsection{Channel Correlation}

Prior to evaluating the aforementioned models, the data from all the channels was analysed to determine whether the hypothesis that the chosen 5 channels are enough to perform EEG-ER has sound foundations. The Pearson correlation coefficient was computed between each of the chosen five channels and the other 31 channels, on the four frequency sub-bands, to analyse whether there is a clear correlation between topologically adjacent channels as opposed to channels that are positioned further away. 

\begin{figure*}
    \centering
    \includegraphics[width=0.9\textwidth]{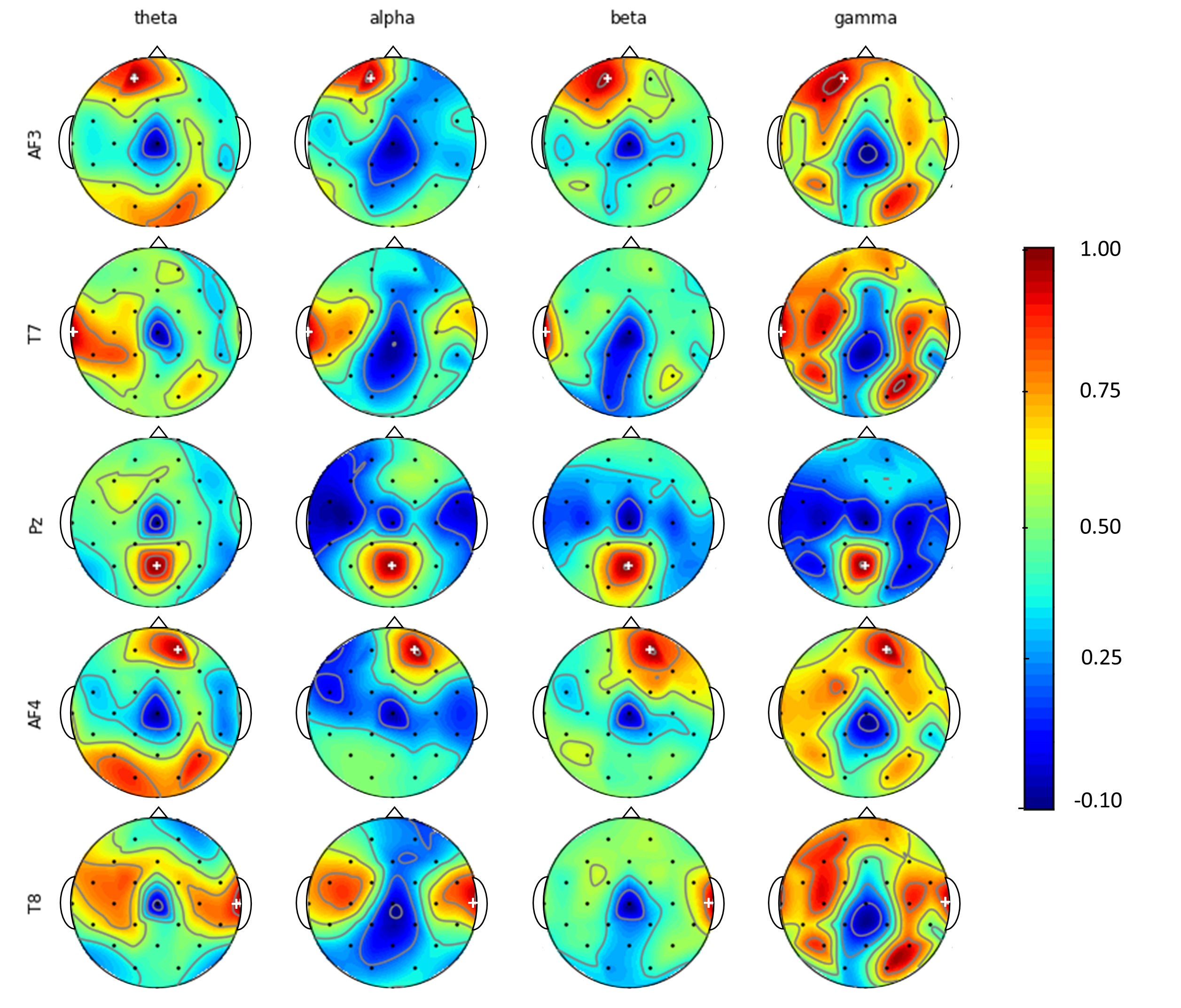}
    \caption{Pearson correlation based on wavelet entropy features between each of the selected 5 channels (indicated with a ``+" sign) and the rest of the channels, on the four frequency bands.}
    \label{fig:corr}
\end{figure*}

Figure \ref{fig:corr} illustrates the correlation between channels, with the respective reference channel marked with a ``+". One can observe that for each of the chosen channels, there is a high correlation with its adjacent channels. This indicates that when using 32 channels, a lot of redundant information is being processed, which might not be needed to achieve high EEG-ER accuracy. 

\subsection{Subject Independent}

\begin{table*}
\centering
\resizebox{\textwidth}{!}{%
\begin{tabular}{|c|c|c|rr|rr|rr|} 
\cline{4-9}
\multicolumn{1}{c}{}      & \multicolumn{1}{c}{}  &                                                                                 & \multicolumn{2}{c|}{Full 32 Channels}                                                                                                     & \multicolumn{2}{c|}{Reduced 5 Channels}                                                                                     & \multicolumn{2}{c|}{\% difference}                                                                         \\ 
\hline
Target                    & Window Size           & Classifier                                                                      & \multicolumn{1}{c}{w/o Baseline}                   & \multicolumn{1}{c|}{Baseline}                                       & \multicolumn{1}{c}{w/o Baseline}                             & \multicolumn{1}{c|}{Baseline}                                & \multicolumn{1}{c}{w/o Baseline}                    & \multicolumn{1}{c|}{Baseline}                        \\ 
\hline \hline
                          & 1                     & 3DCNN (DE) \cite{Yang2018ContinuousRecognition}                                                                     & 55.31 ($\pm$5.53)                                                              & 55.31 ($\pm$5.53)                                                               & 55.31 ($\pm$4.97)                                                       & 55.31 ($\pm$4.96)                                                       & 0.00                                                & 0.00                                                 \\
                          &                       & 3DCNN (DWT)                                                                     & 55.31 ($\pm$5.52)                                                              & 55.31 ($\pm$5.52)                                                              & 55.31 ($\pm$4.15)                                                        & 55.31 ($\pm$4.16)                                                        & 0.00                                                & 0.00                                                 \\
\multicolumn{1}{|l|}{}    & \multicolumn{1}{l|}{} & \multicolumn{1}{l|}{3DCNN (Stacked)}                                            & 55.31 ($\pm$4.95)                                                              & 55.31 ($\pm$4.95)                                                              & 55.31 ($\pm$4.32)                                                       & 55.31 ($\pm$4.32)                                                       & 0.00                                                & 0.00                                                 \\
                          &                       & SVM                                                                             & \textbf{64.36 ($\pm$3.1)}                                                     & \textbf{83.79 ($\pm$2.2)}                                             & 63.36 ($\pm$3.8)                                                        & 79.58  ($\pm$1.4)                                                       & -1.55                                               & -5.02                                                \\
\multicolumn{1}{|l|}{VAL} & \multicolumn{1}{l|}{} & \multicolumn{1}{l|}{{\cellcolor[rgb]{0.753,0.753,0.753}}\textit{SVM (VALARO) }} & {\cellcolor[rgb]{0.753,0.753,0.753}}\textbf{\textit{68.0 ($\pm$5.3)}}         & {\cellcolor[rgb]{0.753,0.753,0.753}}\textbf{\textit{84.0 ($\pm$1.8)}}          & {\cellcolor[rgb]{0.753,0.753,0.753}}\textit{64.34 ($\pm$3.6)}          & {\cellcolor[rgb]{0.753,0.753,0.753}}\textit{81.68 ($\pm$2.6)}           & {\cellcolor[rgb]{0.753,0.753,0.753}}\textit{-5.00 } & {\cellcolor[rgb]{0.753,0.753,0.753}}\textit{-2.76 }  \\ 
\cline{2-9}
                          & 3                     & 3DCNN (DE) \cite{Yang2018ContinuousRecognition}                                                                    & 55.31 ($\pm$3.53)                                                              & 55.31 ($\pm$3.53)                                                               & 55.31  ($\pm$5.01)                                                      & 55.31 ($\pm$3.87)                                                       & 0.00                                                & 0.00                                                 \\
                          &                       & 3DCNN (DWT)                                                                     & 55.31 ($\pm$3.53)                                                              & 55.31 ($\pm$3.53)                                                               & 55.31 ($\pm$4.95)                                                       & 55.31 ($\pm$4.92)                                                        & 0.00                                                & 0.00                                                 \\
                          &                       & 3DCNN (Stacked)                                                                 & 55.31 ($\pm$4.82)                                                               & 55.31 ($\pm$4.82)                                                              & 55.31  ($\pm$4.82)                                                       & 55.31  ($\pm$4.83)                                                      & 0.00                                                & 0.00                                                 \\
                          &                       & SVM                                                                             & 63.10 ($\pm$6.8)                                                               & 82.45 ($\pm$4.2)                                                               & \textbf{64.86 ($\pm$6.4)}                                              & \textbf{80.70 ($\pm$6.8)}                                                 & 2.79                                                & -2.12                                                \\
\multicolumn{1}{|l|}{}    & \multicolumn{1}{l|}{} & \multicolumn{1}{l|}{{\cellcolor[rgb]{0.753,0.753,0.753}}\textit{SVM (VALARO) }} & {\cellcolor[rgb]{0.753,0.753,0.753}}\textit{67.0 ($\pm$7.12)}                  & {\cellcolor[rgb]{0.753,0.753,0.753}}\textit{83.0 ($\pm$2.71)}                  & {\cellcolor[rgb]{0.753,0.753,0.753}}\textbf{\textit{67.24 ($\pm$5.01)}} & {\cellcolor[rgb]{0.753,0.753,0.753}}\textbf{\textit{82.75 ($\pm$9.43)}} & {\cellcolor[rgb]{0.753,0.753,0.753}}\textit{0.35 }  & {\cellcolor[rgb]{0.753,0.753,0.753}}\textit{0.3 }    \\ 
\hline \hline
                          & 1                     & 3DCNN (DE) \cite{Yang2018ContinuousRecognition}                                                                     & 57.58 ($\pm$4.18)                                                                  & 57.58 ($\pm$4.18)                                                                  & 57.58 ($\pm$4.34)                                                         & 57.58 ($\pm$4.34)                                                           & 0.00                                                & 0.00                                                 \\
                          &                       & 3DCNN (DWT)                                                                     & 57.58  ($\pm$4.18)                                                                 & 57.58 ($\pm$4.18)                                                                & 57.58  ($\pm$3.47)                                                          & 60.24 ($\pm$4.34)                                                        & 0.00                                                & 4.62                                                 \\
                          &                       & 3DCNN (Stacked)                                                                 & 57.58 ($\pm$4.34)                                                              & 57.58 ($\pm$3.16)                                                               & 57.58 ($\pm$3.78)                                                        & 57.58 ($\pm$4.33)                                                      & 0.00                                                & 0.00                                                 \\
                          &                       & SVM                                                                             & \textbf{66.90 ($\pm$2.4)}                                                     & \textbf{84.31 ($\pm$2.5)}                                                      & 64.34 ($\pm$3.1)                                                       & 80.54 ($\pm$3.9)                                                        & -2.49                                               & -3.96                                                \\
\multicolumn{1}{|l|}{ARO} & \multicolumn{1}{l|}{} & \multicolumn{1}{l|}{{\cellcolor[rgb]{0.753,0.753,0.753}}\textit{SVM (AROVAL) }} & {\cellcolor[rgb]{0.753,0.753,0.753}}\textbf{\textbf{\textit{70.0 ($\pm$1.81)}}} & {\cellcolor[rgb]{0.753,0.753,0.753}}\textbf{\textbf{\textit{85.0 ($\pm$2.41)}}} & {\cellcolor[rgb]{0.753,0.753,0.753}}\textit{66.86 ($\pm$3.0)}          & {\cellcolor[rgb]{0.753,0.753,0.753}}\textit{82.57 ($\pm$2.9)}          & {\cellcolor[rgb]{0.753,0.753,0.753}}\textit{-4.49 } & {\cellcolor[rgb]{0.753,0.753,0.753}}\textit{-2.85 }  \\ 
\cline{2-9}
                          & 3                     & 3DCNN (DE) \cite{Yang2018ContinuousRecognition}                                                                     & 57.58 ($\pm$3.85)                                                              & 57.58 ($\pm$3.85)                                                               & 56.68 ($\pm$4.01)                                                        & 55.49 ($\pm$2.80)                                                       & -1.56                                               & -3.63                                                \\
                          &                       & 3DCNN (DWT)                                                                     & 57.58 ($\pm$3.85)                                                               & 57.58 ($\pm$3.84)                                                              & 58.68 ($\pm$3.47)                                                        & 57.58 ($\pm$4.07)                                                        & 1.91                                                & 0.00                                                 \\
                          &                       & 3DCNN (Stacked)                                                                 & 57.58 ($\pm$3.67)                                                              & 57.58  ($\pm$3.67)                                                               & 57.58 ($\pm$3.67)                                                       & 57.58 ($\pm$3.67)                                                       & -1.64                                               & 0.00                                                 \\
                          &                       & SVM                                                                             & 65.98 ($\pm$7.21)                                                              & 83.86 ($\pm$3.83)                                                               & \textbf{65.96 ($\pm$8.31)}                                              & \textbf{81.41 ($\pm$9.10)}                                               & -0.03                                               & -2.92                                                \\
\multicolumn{1}{|l|}{}    & \multicolumn{1}{l|}{} & \multicolumn{1}{l|}{{\cellcolor[rgb]{0.753,0.753,0.753}}\textit{SVM (AROVAL) }} & {\cellcolor[rgb]{0.753,0.753,0.753}}\textit{69.0 ($\pm$5.7)}                  & {\cellcolor[rgb]{0.753,0.753,0.753}}\textit{84.0 ($\pm$4.5) }                    & {\cellcolor[rgb]{0.753,0.753,0.753}}\textbf{\textit{68.42 ($\pm$6.2)}} & {\cellcolor[rgb]{0.753,0.753,0.753}}\textbf{\textit{83.44 ($\pm$5.9)}} & {\cellcolor[rgb]{0.753,0.753,0.753}}\textit{-0.84 } & {\cellcolor[rgb]{0.753,0.753,0.753}}\textit{-0.67 }  \\
\hline
\end{tabular}
}
\caption{Mean subject-independent accuracy results (\%) together with standard deviation, for Valence and Arousal across 1s and 3s time-window sizes, with or without baseline removal, for both 32-channel and reduced 5-channel setups.}
\label{tab:subjindept}
\end{table*}

Table \ref{tab:subjindept} shows the accuracy results for the evaluated subject-independent models. In all scenarios the SVM had a superior performance to the 3DCNN, which had a bad performance overall irrespective of whether the baseline removal technique was used or not. For the full 32-channel setup, a time-window of 1 second was sufficient to achieve the best performance for both Valence and Arousal, while for the reduced 5-channel setup, a time-window of 3 seconds achieved a slightly superior accuracy, albeit in both cases the difference was very small. The best accuracy for the full 32-channel was achieved using the baseline removal technique, achieving 83.79\% for Valence and 84.31\% for Arousal. Similarly, the reduced 5-channel also had its best results using the baseline removal technique, with 80.7\% accuracy for Valence and 81.41\% accuracy for Arousal. Interestingly, the chained SVM models VALARO and AROVAL consistently improved this accuracy even more, achieving 84.0\% for Valence and 85.0\% for Arousal on the 32-channel setup, and 82.75\% for Valence and 83.44\% for Arousal. However, one must keep in mind that these models need to be paired with their SVM counterparts to work, i.e. SVM (VALARO) needs to be paired with the ARO SVM, while SVM (AROVAL) needs to be paired with the VAL SVM. The loss of accuracy between the full 32-channel and reduced 5-channel setups was also consistently low, confirming that EEG-ER can still be achieved with reliable accuracy with only a few channels. 

\subsection{Subject Dependent}

\begin{table*}
\centering
\resizebox{\textwidth}{!}{%
\begin{tabular}{|c|c|c|rr|rr|rr|} 
\cline{4-9}
\multicolumn{1}{c}{}   & \multicolumn{1}{c}{}  &                                                                                & \multicolumn{2}{c|}{Full 32 Channels}                                                                                                     & \multicolumn{2}{c|}{Reduced 5 Channels}                                                                                                   & \multicolumn{2}{c|}{\% difference}                                                                       \\ 
\hline
Target                 & Window Size           & Classifier                                                                     & \multicolumn{1}{c}{w/o Baseline}                                    & \multicolumn{1}{c|}{Baseline}                                       & \multicolumn{1}{c}{w/o Baseline}                                    & \multicolumn{1}{c|}{Baseline}                                       & \multicolumn{1}{c}{w/o Baseline}                   & \multicolumn{1}{c|}{Baseline}                       \\ 
\hline \hline
                       & 1                     & 3DCNN (DE) \cite{Yang2018ContinuousRecognition}                                                                    & \textbf{68.52 ($\pm$6.03)}                                              & 89.45 ($\pm$4.42)                                                       & 63.66 ($\pm$6.95)                                                       & 73.20 ($\pm$3.80)                                                       & -7.09                                              & -18.17                                              \\
                       &                       & 3DCNN (DWT)                                                                    & 62.80 ($\pm$7.14)                                                       & 82.83 ($\pm$9.02)                                                       & 65.96 ($\pm$6.98)                                                       & 82.45 ($\pm$4.18)                                                       & 5.03                                               & -0.46                                               \\
                       &                       & 3DCNN (Stacked)                                                                & 66.63 ($\pm$7.32)                                                       & 88.75 ($\pm$6.93)                                                       & 64.95 ($\pm$7.02)                                                       & 84.17 ($\pm$7.41)                                                       & -2.52                                              & -5.16                                               \\
VAL                    &                       & SVM                                                                            & 67.24 ($\pm$5.68)                                                       & 94.41 ($\pm$3.02)                                                       & 67.45 ($\pm$6.64)                                                       & 90.59 ($\pm$4.31)                                                       & 0.31                                               & -4.05                                               \\
\multicolumn{1}{|l|}{} & \multicolumn{1}{l|}{} & \multicolumn{1}{l|}{{\cellcolor[rgb]{0.753,0.753,0.753}}\textit{SVM (VALARO)}} & {\cellcolor[rgb]{0.753,0.753,0.753}}\textit{71.97 ($\pm$5.61)}          & {\cellcolor[rgb]{0.753,0.753,0.753}}\textit{95.26 ($\pm$2.6)}           & {\cellcolor[rgb]{0.753,0.753,0.753}}\textbf{\textit{72.54 ($\pm$6.86)}} & {\cellcolor[rgb]{0.753,0.753,0.753}}92.45 ($\pm$3.92)                   & {\cellcolor[rgb]{0.753,0.753,0.753}}\textit{0.79}  & {\cellcolor[rgb]{0.753,0.753,0.753}}\textit{-2.94}  \\ 
\cline{2-9}
                       & 3                     & 3DCNN (DE) \cite{Yang2018ContinuousRecognition}                                                                     & 65.19 ($\pm$6.28)                                                       & 82.31 ($\pm$5.02)                                                       & 62.16 ($\pm$6.43)                                                       & 72.48 ($\pm$4.06)                                                       & -4.65                                              & -11.94                                              \\
                       &                       & 3DCNN (DWT)                                                                    & 63.67 ($\pm$6.80)                                                       & 81.63 ($\pm$9.66)                                                       & \textbf{68.01 ($\pm$6.54) }                                             & 86.18 ($\pm$3.87)                                                       & 6.82                                               & 5.57                                                \\
                       &                       & 3DCNN (Stacked)                                                                & 67.35 ($\pm$6.64)                                                       & 86.51 ($\pm$7.86)                                                       & 67.91 ($\pm$6.38)                                                       & 91.38 ($\pm$2.72)                                                       & 0.83                                               & 5.63                                                \\
                       &                       & SVM                                                                            & 68.10 ($\pm$5.55)                                                       & \textbf{\textbf{95.32 ($\pm$2.99)}}                                     & 66.99 ($\pm$6.83)                                                       & \textbf{92.2 ($\pm$4.41)}                                               & -1.63                                              & -3.27                                               \\
\multicolumn{1}{|l|}{} & \multicolumn{1}{l|}{} & \multicolumn{1}{l|}{{\cellcolor[rgb]{0.753,0.753,0.753}}\textit{SVM (VALARO)}} & {\cellcolor[rgb]{0.753,0.753,0.753}}\textbf{\textit{72.37 ($\pm$5.71)}} & {\cellcolor[rgb]{0.753,0.753,0.753}}\textit{\textbf{95.93 ($\pm$2.59)}} & {\cellcolor[rgb]{0.753,0.753,0.753}}\textit{72.15 ($\pm$7.47)}          & {\cellcolor[rgb]{0.753,0.753,0.753}}\textbf{\textit{93.81 ($\pm$3.77)}} & {\cellcolor[rgb]{0.753,0.753,0.753}}\textit{-0.3}  & {\cellcolor[rgb]{0.753,0.753,0.753}}\textit{-2.21}  \\ 
\hline \hline 
                       & 1                     & 3DCNN (DE) \cite{Yang2018ContinuousRecognition}                                                                    & \textbf{69.55 ($\pm$7.46)}                                              & 90.24 ($\pm$4.08)                                                       & 66.26 ($\pm$9.16)                                                       & 76.52 ($\pm$4.83)                                                       & -4.73                                              & -15.20                                              \\
                       &                       & 3DCNN (DWT)                                                                    & 64.74 ($\pm$9.10)                                                       & 84.01 ($\pm$9.29)                                                       & 67.99 ($\pm$8.98)                                                       & 83.58 ($\pm$4.37)                                                       & 5.02                                               & -0.51                                               \\
                       &                       & 3DCNN (Stacked)                                                                & 67.56 ($\pm$7.91)                                                       & 89.74 ($\pm$7.45)                                                       & 66.22 ($\pm$9.02)                                                       & 85.62 ($\pm$6.77)                                                       & -1.98                                              & -4.59                                               \\
ARO                    &                       & SVM                                                                            & 68.92 ($\pm$6.86)                                                       & 94.86 ($\pm$2.96)                                                       & 67.90 ($\pm$6.88)                                                       & 90.97 ($\pm$3.60)                                                       & -1.48                                              & -4.10                                               \\
\multicolumn{1}{|l|}{} & \multicolumn{1}{l|}{} & \multicolumn{1}{l|}{{\cellcolor[rgb]{0.753,0.753,0.753}}\textit{SVM (AROVAL)}} & {\cellcolor[rgb]{0.753,0.753,0.753}}\textit{\textbf{73.37 (±5.95)}} & {\cellcolor[rgb]{0.753,0.753,0.753}}\textit{95.58 (±2.58)}          & {\cellcolor[rgb]{0.753,0.753,0.753}}\textit{\textbf{72.58 (±6.27)}} & {\cellcolor[rgb]{0.753,0.753,0.753}}\textit{92.91 (±3.19)}          & {\cellcolor[rgb]{0.753,0.753,0.753}}\textit{-1.08} & {\cellcolor[rgb]{0.753,0.753,0.753}}\textit{-2.79}  \\ 
\cline{2-9}
                       & 3                     & 3DCNN (DE) \cite{Yang2018ContinuousRecognition}                                                                     & 66.59 (±7.66)                                                       & 83.9 (±4.43)                                                        & 65.11 (±9.04)                                                       & 75.29 (±5.48)                                                       & -2.22                                              & -10.26                                              \\
                       &                       & 3DCNN (DWT)                                                                    & 64.82 (±8.95)                                                       & 83.1 (±8.99)                                                        & 68.78 (±8.49)                                                       & 86.88 (±4.27)                                                       & 6.11                                               & 4.55                                                \\
                       &                       & 3DCNN (Stacked)                                                                & 67.75 (±7.44)                                                       & 87.69 (±6.67)                                                       & \textbf{69.51 (±8.08) }                                             & 91.81 (±2.89)                                                       & 2.60                                               & 4.70                                                \\
                       &                       & SVM                                                                            & 69.05 (±7.23)                                                       & \textbf{\textbf{95.68 (±2.87)}}                                     & 67.50 (±6.83)                                                       & \textbf{92.13 (±3.53)}                                              & -2.24                                              & -3.71                                               \\
\multicolumn{1}{|l|}{} & \multicolumn{1}{l|}{} & \multicolumn{1}{l|}{{\cellcolor[rgb]{0.753,0.753,0.753}}\textit{SVM (AROVAL)}} & {\cellcolor[rgb]{0.753,0.753,0.753}}\textit{73.22 (±6.41)}          & {\cellcolor[rgb]{0.753,0.753,0.753}}\textit{\textbf{96.26 (±2.31)}} & {\cellcolor[rgb]{0.753,0.753,0.753}}\textit{72.12 (±6.76)}          & {\cellcolor[rgb]{0.753,0.753,0.753}}\textit{\textbf{94.19 (±2.92)}} & {\cellcolor[rgb]{0.753,0.753,0.753}}\textit{-1.5}  & {\cellcolor[rgb]{0.753,0.753,0.753}}\textit{-2.15}  \\
\hline
\end{tabular}
}
\caption{Mean subject-dependent accuracy results (\%) together with standard deviation, for Valence and Arousal across 1s and 3s time-windows, with or without baseline removal, for both 32-channel and reduced 5-channel setups.}
\label{tab:subjdept}
\end{table*}

Table \ref{tab:subjdept} shows the results for the mean subject-dependent accuracies across all subjects, together with the standard deviation. Once again, there is a significant difference in performance between the models that use baseline removal preprocessing and those that do not. The 3DCNN using Differential Entropy \cite{Yang2018ContinuousRecognition} had the best accuracy when baseline removal was not used for the full 32-channel case, at 68.52\% for Valence and 69.55\% for Arousal using a 1 second time-window. This was followed closely by the SVM model with 68.10\% for Valence and 69.05\% for Arousal using a 3-second time window. However, when using the baseline removal preprocessing, the accuracy of the SVM model went up to 95.32\% for Valence and 95.68\% for Arousal, surpassing the current state-of-the-art 3DCNN DE model by more than 6\%. 

On the reduced 5-channel setup, the best accuracy without baseline removal preprocessing for Valence was obtained with our proposed 3DCNN with DWT Entropy, at 68.01\% using a 3-second time window, together with our proposed 3DCNN with Stacked DWT Entropy and DE achieving 69.51\% for Arousal. Once again, when using the baseline removal preprocessing, the SVM surpassed all models, achieving 92.2\% accuracy on Valence and 92.13\% or Arousal, surpassing the 3DCNN DE model by more than 25\% on Valence and more than 20\% on Arousal. 

Similarly to the subject-independent case, the chained SVM models pushed up the accuracy higher, surpassing every model on both the full 32-channel and reduced 5-channel scenarios, with or without baseline removal. However, one must keep in mind that only one of the SVM (VALARO) or SVM (AROVAL) models can be used concurrently, since they have to be paired with their corresponding ARO SVM or VAL SVM respectively. 

The drop in accuracy between the 32 channels and 5 channels was less than 4\% for the best performing models using baseline removal preprocessing, while when this was not used, the drop in accuracy was of 7\% on Valence and 5\% on Arousal for the best performing model (3DCNN DE \cite{Yang2018ContinuousRecognition}). This was brought down to just around 1\% when using the chained SVM models. 

\begin{figure*}
\centering
\begin{subfigure}[b]{0.5\textwidth}
  \centering
  \includegraphics[width=\textwidth]{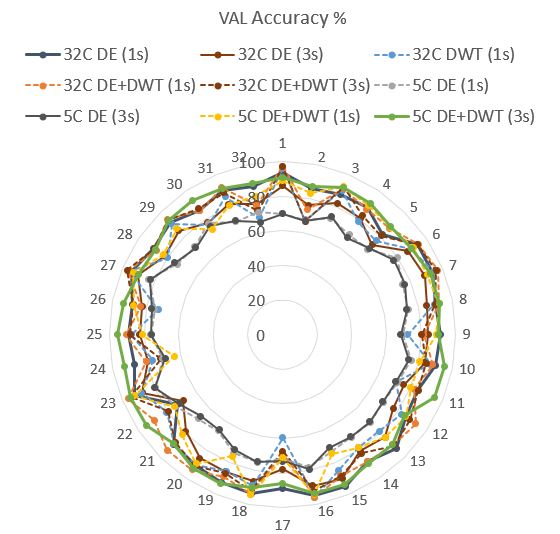}
  \caption{Valence}
\end{subfigure}%
\begin{subfigure}[b]{0.5\textwidth}
  \centering
  \includegraphics[width=\textwidth]{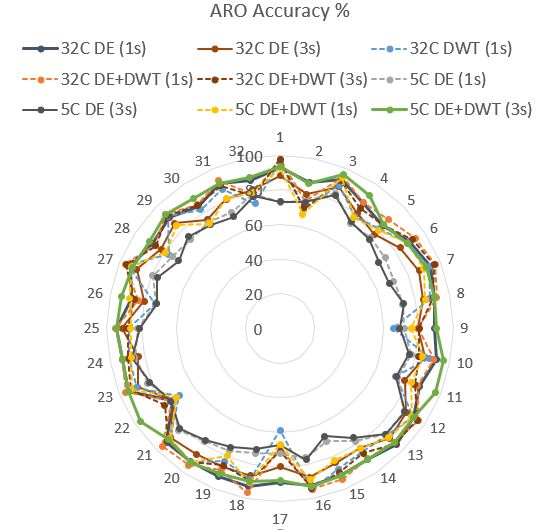}
  \caption{Arousal}
\end{subfigure}
\caption{Comparison of subject-dependent 3DCNN accuracies on each of the 32 subjects in the DEAP dataset \cite{KoelstraDEAP:Signals} using baseline removal preprocessing, on both 32 channels and 5 channels, including DWT Entropy and/or DE features with $\tau \in \{1,3\}$.}
\label{fig:32v5_features_dept_radar}
\end{figure*}

An interesting observation is that the 3DCNN with stacked DWT Entropy and DE for the  5-channel case when using the baseline removal technique and using 3 second time windows had a higher accuracy than the 3DCNN with DE \cite{Yang2018ContinuousRecognition} and the 3DCNN with DWT Entropy and DE for the full 32-channels with 1 second time windows. While both use a very similar 3DCNN architecture with the same number of layers and configuration, the addition of DWT Entropy with a larger time window yielded a better performance. Figure \ref{fig:32v5_features_dept_radar} shows the accuracy for each of the 32 subjects in the dataset, for the 32-channel 3DCNN DE model $(\tau = 1)$ \cite{Yang2018ContinuousRecognition}, together with other 3DCNN 32-channel and 5-channel variants. The 5-channel DE+DWT model $(\tau = 3)$ maintains the highest and most consistent accuracy across all subjects.  

\begin{figure*}
\centering
\begin{subfigure}[b]{0.5\textwidth}
  \centering
  \includegraphics[width=0.75\textwidth]{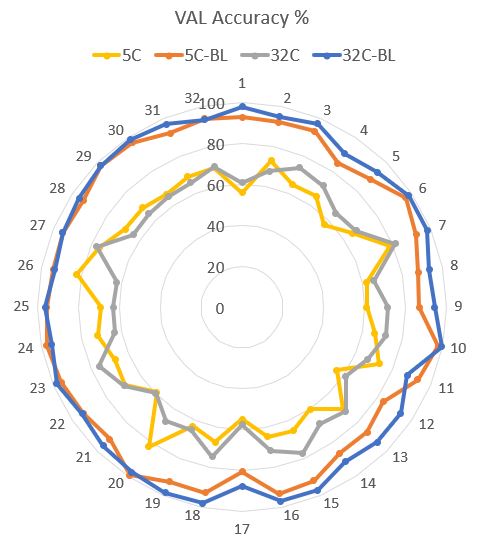}
  \caption{Valence}
\end{subfigure}%
\begin{subfigure}[b]{0.5\textwidth}
  \centering
  \includegraphics[width=0.75\textwidth]{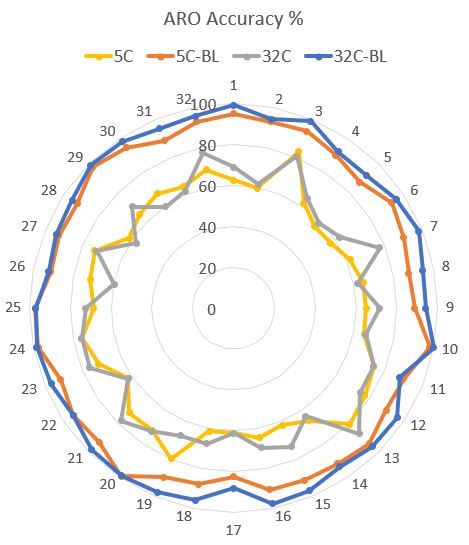}
  \caption{Arousal}
\end{subfigure}
\caption{Comparison of the subject-dependent SVM accuracy $(\tau = 3)$ on each of the 32 subjects in the DEAP dataset \cite{KoelstraDEAP:Signals} with and without using baseline removal preprocessing \cite{Yang2018ContinuousRecognition, Tao2020EEG-basedAttention}, for both full 32-channel and reduced 5-channel scenarios.}
\label{fig:baselinereduction}
\end{figure*}

As shown in Figure \ref{fig:baselinereduction} the difference in classification accuracy between the full 32-channels and the reduced 5-channel setup is not significant, sometimes even having an inverse effect of increasing accuracy due to less model dimensionality and complexity. What makes a significant difference is whether the baseline removal technique is used, bringing the accuracy of both Valence and Arousal above 80\% for all subjects in the dataset. 

\section{Discussion}

\begin{figure}
    \centering
    \includegraphics[width=0.48\textwidth]{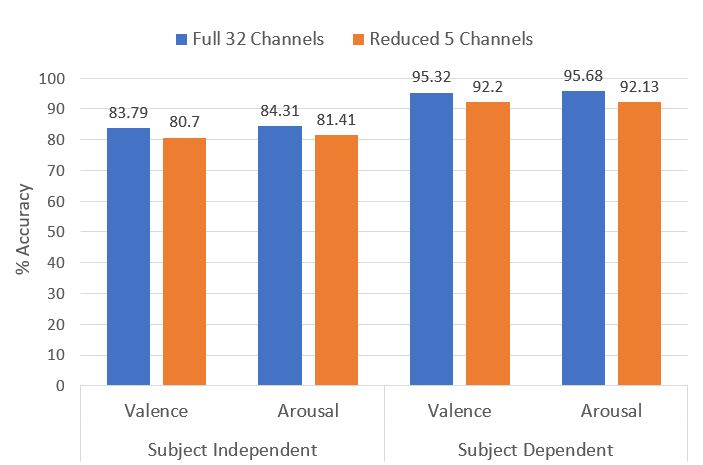}
    \caption{Comparison of the best accuracy for Valence and Arousal. These were all achieved by the proposed SVM model.}
    \label{fig:accuracychart}
\end{figure}

The objective of this research was to find the best models for real-time EEG-ER that perform well in both full 32-channel and reduced 5-channel modes. Figure \ref{fig:accuracychart} shows the best results for both subject-independent and subject-dependent cases. These were all achieved using the proposed SVM with RBF kernel model that makes use of the DWT Entropy and Energy features with baseline removal preprocessing. The mean drop in accuracy between 32-channels and 5-channels is only 3.53\%, showing that having a few strategically located EEG sensors is enough to perform reliable EEG-ER. Furthermore, the model performs well in both subject-independent and subject-dependent cases. The chained SVM architecture also offers the opportunity to increase the accuracy of one of the dimensions, Valence or Arousal, a few percentage points higher, by using the output of one classifier as an extra input to the other classifier. The choice of time window size does not seem to have a significant effect on the SVM classification accuracy, although the 3-second time window performed marginally better in most cases.

Apart from achieving a high accuracy for reduced 5-channel scenarios, the results on the full 32-channels of the DEAP \cite{KoelstraDEAP:Signals} dataset beat the current state-of-the-art accuracy achieved by subject-dependent models. Our proposed SVM has the highest accuracy, 95.3\% for Valence and 95.7\% for Arousal, while the best accuracy that was achieved at the time of writing was of 92.7\% for Valence and 93.1\% for Arousal using an ACRNN model \cite{Tao2020EEG-basedAttention}. 

\section{Conclusions \& Future Work}

This work proposes a real-time solution for EEG-ER using a reduced number of channels, thus opening up the possibilities to make use of less costly consumer-grade EEG devices. A time-windowing approach is used to classify a person's emotion from a few seconds of data, thus providing the opportunity to create BCI systems that interact with users in response to their current emotional state. 

The DEAP \cite{KoelstraDEAP:Signals} dataset was used to benchmark the classification models and compare them with those described in prior work. We propose to use DWT to extract time-frequency domain features for both subject-independent models, where one generic model works across all individuals, and subject-dependent models, where the data from each participant is used to create a personalised model for each individual.  We also adapt the baseline removal preprocessing technique \cite{Yang2018ContinuousRecognition,Tao2020EEG-basedAttention}, that computes the difference between the individual's active state and rest state, for the DWT features. This preprocessing technique delivers a significant increase in accuracy across both subject-independent and subject-dependent models. 

The 3D CNN model that makes use of DE \cite{Yang2018ContinuousRecognition} was adapted to work with DWT Entropy and also a combination of both DE and DWT Entropy features. However, an SVM with RBF kernel, making use of DWT Entropy and Energy features proved to perform best on both full 32-channel and reduced 5-channel scenarios, beating the known state-of-the-art benchmarks. Furthermore, the mean difference in accuracy between 32 channels and 5 channels for the SVM model is only 3.54\%, showing that performing accurate EEG-ER with devices that only have a few channels is feasible. 

We also propose a chained SVM model, where the Arousal and Valence classifiers are chained together, with the output of the Arousal SVM feeding into the Valence SVM, or vice-versa. This approach takes advantage of the conditional independence of both Arousal and Valence on the same emotional state, where knowing one of the dimensions can influence the interpretation of the data to predict the other dimension. These models achieve a slightly higher accuracy than the single SVM model.

Future work includes validating these models on other datasets, such as SEED \cite{duan2013differential,Zheng2015InvestigatingNetworks}, DREAMER  \cite{katsigiannis2017dreamer}, and MAHNOB-HCI \cite{soleymani2011multimodal}. This can then be followed by real-world experiments using the actual consumer-grade devices, such as the Emotiv Insight, in order to assess the performance and generality of the models. Other dimensions of emotion, such as Dominance, can also be explored. The effect of further data preprocessing the data, such as using Independent Component Analysis (ICA) or Principal Component Analysis can also be investigated, which could lead to lower dimensionality (especially in the 32 channel case) and better convergence during training of the classifiers.

Furthermore, other classification techniques can also be investigated, such as the ones proposed very recently to detect epileptic seizures \cite{tziridis2021eeg,zeng2021grp}, which make use of Recurrence Plots \cite{eckmann1995recurrence,marwan2007recurrence}. 

Achieving a high EEG-ER accuracy with affordable consumer-grade EEG devices opens up a lot of opportunities for applications. BCI-based emotion-adaptive entertainment systems could recommend music or movies according to the user's mood. Video games could also adapt the environment or intensity of the game according to the player's real-time emotion state. Real-time EEG-ER can also be useful to diagnose mental or mood disorders, or even during psychiatric therapy sessions, providing the practitioner with additional real-time information about the feelings of the patient. The machine learning models presented in this work can be further developed and integrated into BCI systems to make these applications a reality. 

\section*{Acknowledgements}

The authors would like to thank the Malta Information Technology Agency (MITA) Emerging Technologies Lab for their support, and the DEAP \cite{KoelstraDEAP:Signals} dataset administrators for granting permissions to use the dataset in this research.

\printcredits

\bibliographystyle{cas-model2-names}

\bibliography{references}




\end{document}